\documentclass[conference]{IEEEtran}
\IEEEoverridecommandlockouts
\usepackage{cite}
\usepackage{amsmath,amssymb,amsfonts}
\usepackage{algorithmic}
\usepackage{graphicx}
\usepackage{textcomp}
\usepackage{xcolor}
\usepackage{multirow}
\usepackage{makecell}

\usepackage{tikz}
\usetikzlibrary{patterns}
\usepackage{pgfplots}
\pgfplotsset{compat=1.18}
\usetikzlibrary{external}%Biblioteca para salvar os tikzs gerados no formato pdf

\def\BibTeX{{\rm B\kern-.05em{\sc i\kern-.025em b}\kern-.08em
    T\kern-.1667em\lower.7ex\hbox{E}\kern-.125emX}}

\usepackage[caption=false]{subfig}
\usepackage[nolist]{acronym}
\usepackage{balance}

\makeatletter
\newcommand{\linebreakand}{%
  \end{@IEEEauthorhalign}
  \hfill\mbox{}\par
  \mbox{}\hfill\begin{@IEEEauthorhalign}
}
\makeatother

\begin{acronym}[TDMA]
    \acro{GPU}{Graphics Processing Unit}
    \acro{CSV}{Comma-Separated Values}
    \acro{CNN}{Convolutional Neural Network}
\end{acronym}

\begin{document}

\title{Optimizing Parking Space Classification: Distilling Ensembles into Lightweight Classifiers\\
\thanks{This work has been supported by the Brazilian National Council for Scientific and Technological Development (CNPq) -- Grant 405511/2022-1, and by the Coordenação de Aperfeiçoamento de Pessoal de Nível Superior – Brasil (CAPES) – Finance Code 001.}
}

\author{
    \IEEEauthorblockN{Paulo Luza Alves\IEEEauthorrefmark{1}, 
        André Hochuli\IEEEauthorrefmark{2},
    Luiz Eduardo de Oliveira\IEEEauthorrefmark{1},
        Paulo Lisboa de Almeida\IEEEauthorrefmark{1}}
    \IEEEauthorblockA{\IEEEauthorrefmark{1}
        Departamento de Informática (DInf),
	Universidade Federal do Paran\'a,
	Curitiba, PR - Brazil\\
        \{paulomateus,luiz.oliveira,paulorla\}@ufpr.br
    }
    \IEEEauthorblockA{\IEEEauthorrefmark{2}
        Programa de Pós-Graduação em Informática (PPGIa),
	Pontif\'icia Universidade Cat\'olica do Paran\'a,
	Curitiba, PR - Brazil\\
	aghochuli@ppgia.pucpr.br
    }
 %    \IEEEauthorblockA{\IEEEauthorrefmark{3}
 %        Department of Informatics (DInf)
	% Universidade Federal do Paran\'a
	% Curitiba, PR - Brazil\\
 %        luiz.oliveira@ufpr.br
 %    }
 %    \IEEEauthorblockA{\IEEEauthorrefmark{4}
 %        Department of Informatics (DInf)
	% Universidade Federal do Paran\'a
	% Curitiba, PR - Brazil\\
 %        paulorla@ufpr.br
 %    }
}

\maketitle

\begin{abstract}
When deploying large-scale machine learning models for smart city applications, such as image-based parking lot monitoring, data often must be sent to a central server to perform classification tasks. This is challenging for the city's infrastructure, where image-based applications require transmitting large volumes of data, necessitating complex network and hardware infrastructures to process the data. To address this issue in image-based parking space classification, we propose creating a robust ensemble of classifiers to serve as Teacher models. These Teacher models are distilled into lightweight and specialized Student models that can be deployed directly on edge devices. The knowledge is distilled to the Student models through pseudo-labeled samples generated by the Teacher model, which are utilized to fine-tune the Student models on the target scenario. Our results show that the Student models, with 26 times fewer parameters than the Teacher models, achieved an average accuracy of 96.6\% on the target test datasets, surpassing the Teacher models, which attained an average accuracy of 95.3\%.
\end{abstract}

% \begin{IEEEkeywords}
% Smart City, Deep Learning, Model Distillation, Parking Space Classification, Edge Computing
% \end{IEEEkeywords}

\section{Introduction}\label{sec:intro}

When deploying machine learning models that deal with image-related tasks, we often face the problem that these models may require specialized hardware, such as \acp{GPU}, to deal with massive amounts of inputs, often requiring the acquired images to be sent to a central server, which has the processing power required.
In a smart city, where thousands of cameras may be deployed to feed the machine learning models, bottlenecks may be created in these servers and the smart city's network infrastructure.

\begin{figure}[htbp]
\centering
\includegraphics[width=0.4\textwidth, trim={1cm 4cm 6cm 5cm},clip]{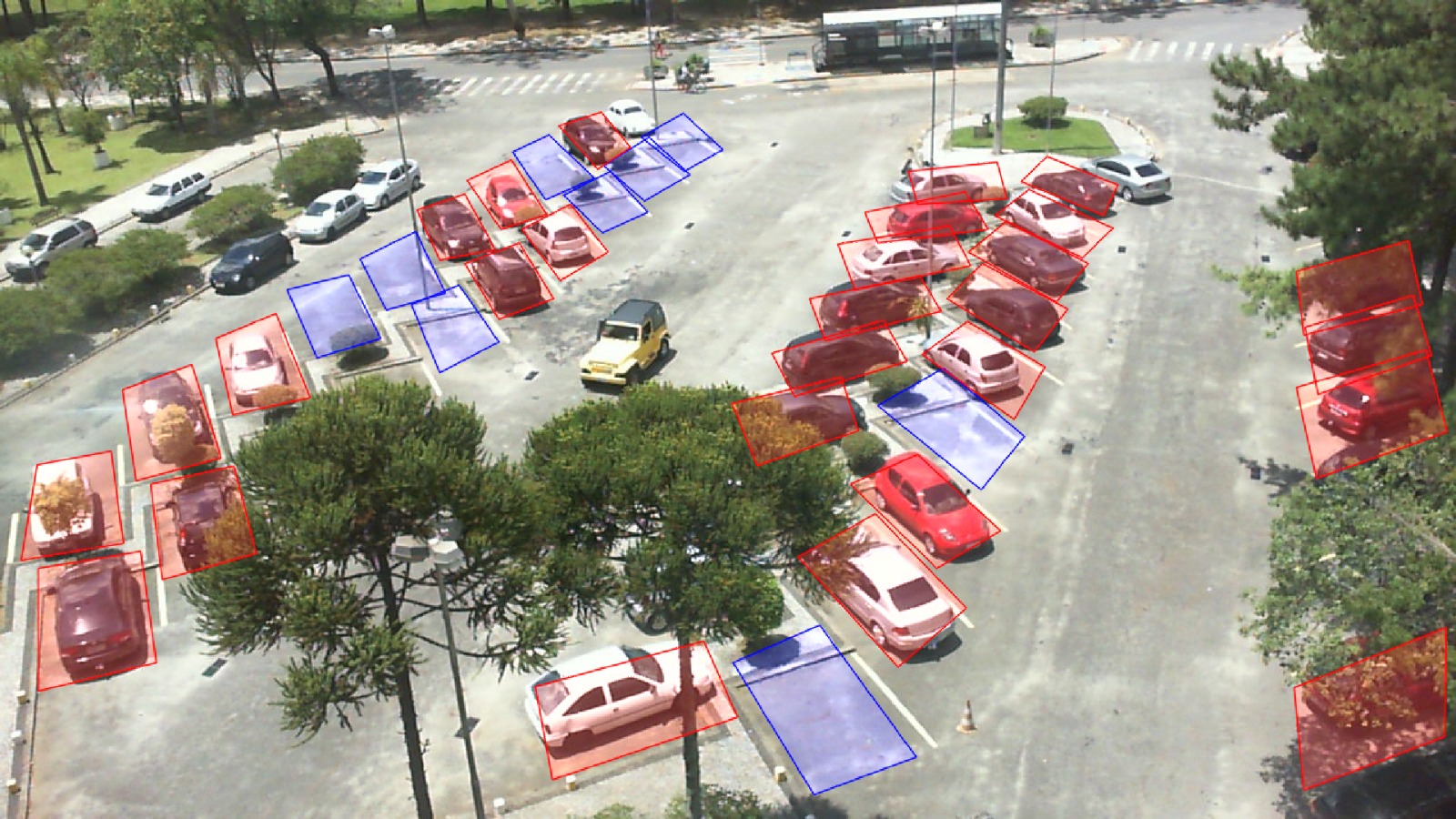}
\caption{Occupied (red) and empty (blue) parking spaces -- PKLot \cite{almeidaEtAl2015}.}
\label{fig:classificationExample}
\end{figure}

We propose distilling parking space classification models on demand (example in Figure \ref{fig:classificationExample}) to create lightweight models that can be processed on edge devices, such as smart cameras. The key idea is to create a robust Teacher Model to classify images from any parking area. This Teacher Model may be computationally expensive and is deployed on a central server. Whenever a new camera is deployed in a parking area, the images of this area are sent and classified by the Teacher model for a short period (e.g., seven days), where the classification results are used as pseudo-labels to train smaller specialist models that can be deployed on the edge.

To provide an understanding of the computational cost for the parking spaces classification problem, consider a relatively small-scale deployment with 1,000 cameras. Suppose each camera sends a $1280 \times 720$ pixels JPEG compressed image to a central server every 30 seconds. In that case, it will result in a transfer of 35 gigabytes of data per hour to the server, not accounting for the network's overheads (estimated using the PKLot dataset~\cite{almeidaEtAl2015}). This data transfer can approximately double for $1920 \times 1080$ pixel images.

%\footnote{Using the PKLot dataset, we estimated that each compressed image, using JPEG with 95\% quality, would require 303 kilobytes of data.}

The hypothesis that small specialist models can be used and deployed in the real world is supported by many works \cite{almeidaEtAl2022,amatoEtAl2017,ahrnbomEtAl2016,almeidaEtAl2015}, although the authors of these lightweight approaches use real labels of the target datasets, creating scalability problems, since every time a new camera is deployed, humans need to label the images from the target parking lot to train the model.

We hypothesize that lightweight models that can be deployed in edge devices, fine-tuned using pseudo-labels, can generate accuracy comparable to a central model but with significantly reduced computational power requirements. To test this hypothesis, we aim to answer the following questions:

\begin{itemize}
    \item RQ1: How does the accuracy of the lightweight Student classifiers compare to the central Teacher model?
    \item RQ2: How many images must be classified by the Teacher model to create enough pseudo-labels to fine-tune the Student models?
    \item RQ3: How does the accuracy of the Student models fine-tuned with the pseudo-labels compare with hypothetical Students fine-tuned with true-labels?
\end{itemize}

As the Teacher models, we employ ensembles of classifiers, where each model in the ensemble is a \ac{CNN} with 4,204,594 parameters. As Student models, we use single \acp{CNN} with 1,519,906 and 158,914 parameters. Our experiments show that the Student models, fine-tuned with pseudo-labeled images, reach over 96\% of accuracy, which is higher than the Teacher models, which reached accuracies of 95.3\%.

% The remainder of this paper is organized as follows: In Section \ref{sec:related_work}, we show the related work. In Section \ref{sec:proposed}, we detail our proposed approach and the \ac{CNN} network architectures used. In Section \ref{sec:protocol}, we present our experimental protocol, while in Section \ref{sec:experiments}, we present and discuss the experiment results. Finally, in Section \ref{sec:conclusion}, we present our conclusions.

\section{Related Work}\label{sec:related_work}

Several state-of-the-art works employ machine learning approaches to classify parking space images as occupied or empty. Notable advancements include the creation of large-scale benchmark datasets, such as PKLot \cite{almeidaEtAl2015} and CNPark-EXT \cite{amatoEtAl2017}, and the development of accurate classifiers using labeled training samples from the target parking lot \cite{almeidaEtAl2013, almeidaEtAl2015, amatoEtAl2017, GrbicKoch2023}. Additionally, lightweight classifiers have been proposed \cite{almeidaEtAl2015, valipourEtAl2016, amatoEtAl2017, hochuli2022evaluation, HochuliEtAl2023}, as well as general classifiers capable of classifying instances from a target parking lot without training samples \cite{almeidaEtAl2015, amatoEtAl2017, HochuliEtAl2023}. For a comprehensive review, see \cite{almeidaEtAl2022}.

Besides the number of publications reporting accuracies over 99\%, the scalability of the proposed approaches is still an open question since 1 -- to reach such high accuracies, the proposed approaches often rely on labeled samples from the target dataset, demanding considerable human labor and; 2 -- approaches able to deal with cross-dataset scenarios, where no samples from the target area are given, often rely on computationally expensive classifiers, needing the images to be sent to a server to be processed. Even so, such approaches rarely exceed 95\% of accuracy~\cite{almeidaEtAl2022}.

For example, \cite{valipourEtAl2016} proposes a lightweight approach using a network with 22,544,384 parameters\footnote{We estimated the number of parameters based on the network's size provided in the paper, which is 86MB, and IEEE 754 single precision values.}. Although this approach can be executed on edge devices, the classifier must be trained with labeled instances from the target dataset, potentially limiting large-scale deployments (e.g., in a smart city). Additionally, the authors demonstrated that, using a Raspberry Pi, the parking statuses can only be refreshed once per minute.

Similarly, the approach proposed in \cite{amatoEtAl2017} generates a lightweight network with approximately 2,686,567 parameters, requiring around 15 seconds to process an image using a Raspberry Pi 2\footnote{It is unclear if the authors accounted for both classification time and overheads, such as the time to crop the images}. More recently, Hochuli et al. \cite{hochuli2022evaluation} proposed a 3-layer convolutional neural network comprising 158,000 parameters. Despite not requiring training instances from the target dataset, this lightweight approach achieved only 80.9\% accuracy with the PKLot and CNRPark-Ext datasets. In comparison, a MobilenetV3 with approximately 4,100,000 parameters achieved 89.9\% accuracy.

To the best of our knowledge, the approach proposed by \cite{molo2024teacherpklot} is the most similar to ours. Their architecture features a large YOLO detector model as the teacher, running on a server, and a lightweight YOLOv5 model implemented as the student. The authors trained the teacher model using a combination of state-of-the-art parking lot datasets, including PKLot \cite{almeidaEtAl2015}. Despite promising results, they only distilled the teacher on the CNRPark-Ext \cite{amatoEtAl2017} dataset. Additionally, the teacher model functions as a detector rather than a classifier and contains 1,760,518 parameters.

Therefore, creating reasonably accurate models that can be deployed on resource-constrained devices without the need for human labeling of samples from the target area remains an open question in the state-of-the-art.

\section{Proposed Approach}\label{sec:proposed}

In this Section, we describe our proposed approach. First, in Section \ref{sec:proposedScheme} we define our proposed scheme to create the Teacher and Student models. In Section \ref{sec:cnnarchs} we describe the \ac{CNN} architectures used in this work.

\subsection{Proposed Scheme}\label{sec:proposedScheme}

Many works show that even relatively simple classifiers can achieve high accuracies for the parking spaces classification problem given that training samples from the target parking lot are given \cite{almeidaEtAl2013,almeidaEtAl2015,amatoEtAl2017,GrbicKoch2023}. Inspired by this, along with the idea of model distillation \cite{alkhulaifi2021knowledge}, we propose the following:

\begin{enumerate}
     \item We train an ensemble of classifiers to reach state-of-the-art results in cross-dataset scenarios, where no training image from the target parking lot is given. The ensemble is the Teacher model and can be computationally expensive, being executed in, for instance, a central server.

    \item When a new parking lot area needs to be monitored, the images are sent to the central server for $n$ days (e.g., 7 days, as in the experiments in Section \ref{sec:fixedDays}). The Teacher will classify the images, and the classification results with \textit{a posteriori} confidences above 0.9 will be used as pseudo-labels.

    \item After the $nth$ day, the pseudo-labels are used to fine-tune a lightweight Student model, which can then be deployed directly on the edge (e.g., in a smart camera). Following the deployment of the Student model, there is no longer a need to send images from the target parking lot to the Teacher model. Instead, only the processed information, such as a CSV file containing the parking space locations and statuses, may be sent to a central server to provide the necessary information to the city's drivers and planners.
    
\end{enumerate}

% The proposed schema is illustrated in Figure \ref{fig:pipeline}. During the initial $n$ days, as shown in Figure \ref{fig:pipeline}a, images from each deployed camera are sent to the central server, where the Teacher model classifies the images to generate relevant information, such as parking space occupancy for city drivers and planners. After $n$ days, once the central server has accumulated enough pseudo-labeled samples, a custom lightweight Student model is trained and deployed on the edge. From then on, the edge device processes the information and sends only the relevant processed data to the central server, as depicted in Figure \ref{fig:pipeline}b.

% \begin{figure}[htbp]
%   \centering
%   \subfloat[Day 1 to day $n$ operation.]{\includegraphics[width=8.0cm]{imgs/Father Son Models Day 1 to n.png}}\hfill
%   \subfloat[After the $nth$ day operation.]{\includegraphics[width=8.0cm]{imgs/Father Son Models Day n and beyond.png}}
%   \caption{Proposed scheme. From day 1 to $n$ (a) the images are sent to the Teacher model to be classified, and the results are stored. After the $nth$ (b) a custom Student model fine-tuned with the pseudo-labels is deployed.}
%   \label{fig:pipeline}
% \end{figure}

\subsection{CNN Architectures}\label{sec:cnnarchs}

The Teacher model comprises an ensemble of the Large version of the MobileNetV3\cite{MobilenetV3} networks. This architecture implements a series of convolutional block improvements, such as residual connections and non-linearity transformations, to reduce computation yet maintain reasonable accuracy in state-of-the-art benchmarks. Each network in the ensemble is trained with a subset of the training set (see details in Section \ref{sec:protocol}). Considering the Teacher model (ensemble), given a test instance $\mathbf{x}$, the predicted class $\hat{y}$ is given by averaging the \textit{a posteriori} probabilities of all networks in the ensemble.

For the Student models, we tested two different architectures. The first is the small version of the MobileNetV3 architecture \cite{MobilenetV3}. The second is a custom compact, cost-effective architecture, investigated in \cite{hochuli2022evaluation, HochuliEtAl2023}, designed to classify parking spots when trained on the target dataset.
This architecture
%depicted in Figure \ref{fig:teacher_model},
consists of three convolutional layers combined with two pooling layers. Classification is performed by a dense layer that concatenates all extracted features. The Student model is based on a single classifier, unlike the Teacher model, which is based on an ensemble. Table \ref{table:cnnsComparison} compares the networks.

% \begin{figure}[htbp]
% \centering
% \includegraphics[width=0.485\textwidth]{./imgs/CNN.png}
% \caption{The Custom Student model comprises 3-convolutional layers to perform feature extraction and a dense layer to perform classification \cite{hochuli2022evaluation}.}\label{fig:model}
% \label{fig:teacher_model}
% \end{figure}

\begin{table}[htpb]
\centering
\caption{Comparison of the architectures used in this work.}
\begin{tabular}{lrr}\hline
Architecture & \# Parameters & Mem. (IEEE 754 s.p.) \\\hline
MobileNetV3 Large \cite{MobilenetV3}& 4,204,594 & 17.00 MB\\
MobileNetV3 Small \cite{MobilenetV3}& 1,519,906 & 6.20 MB\\
Custom \cite{hochuli2022evaluation, HochuliEtAl2023} & 158,914 & 0,64 MB\\\hline
\end{tabular}
\label{table:cnnsComparison}
\end{table}

\section{Experimental Protocol}\label{sec:protocol}

Our experiments use the PKLot~\cite{almeidaEtAl2015} and CNRPark-EXT~\cite{amatoEtAl2017} datasets. Table \ref{table:datasets} presents the key properties of these datasets. In total, we used 1,338,879 samples for our experiments\footnote{The total number of samples exceeds those published in the original works \cite{almeidaEtAl2015, amatoEtAl2017} due to the inclusion of newly labeled samples. Additionally, we standardized annotations as rotated rectangles for all parking lots.}. The PKLot dataset includes two distinct parking lots: PUCPR and UFPR. There is a single camera angle for the PUCPR parking lot, while there are two different camera angles for the UFPR, each capturing images on separate days. The CNRPark-EXT dataset consists of nine camera angles simultaneously capturing images from different parking lot sections on the same set of days. In the PKLot dataset, one image is taken every 5 minutes, while in the CNRPark-EXT dataset, there is one image every 30 minutes (for every camera angle).

\begin{table}[htpb]
\centering
\caption{Datasets used in the experiments.}
\setlength{\tabcolsep}{5.1pt}
\begin{tabular}{rrrrrr}\hline
\multicolumn{6}{c}{PKLot -- 12,171 images}            \\\hline
\# Days & \# Park. Lots & \# Angles & \# Occupied & \# Empty & \# Total \\
    100    &   2         &    3      & 543,436 & 648,232 & 1,191,668\\\hline
\multicolumn{6}{c}{CNRPark-EXT -- 4,073 images}            \\\hline
\# Days & \# Park. Lots & \# Angles & \# Occupied & \# Empty & \# Total \\
    23    &   1         &    9      &  81,062 & 66,149 & 147,211\\\hline
\end{tabular}
\label{table:datasets}
\end{table}

We employ one dataset for training the Teacher models and the other for deploying and testing the Student models in a cross-dataset scenario \cite{almeidaEtAl2022}. Specifically, we conduct tests where PKLot serves as the training dataset and CNRPark-EXT as the testing dataset, and vice versa. The following strategy was adopted to train the Teacher model ensemble:

\begin{itemize}
    \item All classifiers are pre-trained in the Imagenet dataset.
    \item We trained a classifier $t_0$ using the initial 70\% of data from each camera angle (ordered chronologically), reserving the remaining data for validation.
    \item For each camera angle $a_i$ available in the dataset, where $i \in [1..k]$ is the index of one of the possible $k$ camera angles, we created a classifier $t_i$ trained with all camera angle images, except $a_i$. We use $a_i$ for validation.
\end{itemize}

Thus, the Teacher model is an ensemble $T$ containing the classifiers $T = [t_0, t_1, \dots , t_k]$. As discussed in Section \ref{sec:proposed}, the images from the first $n$ days of the test (target) parking lot are classified using the Teacher model $T$. Then, the classification results are used as pseudo-labels to train the Student models. We create one student model for each camera angle. Each Student model is initially trained using the same training dataset employed to create $t_0$ and then fine-tuned using images specific to its camera angle and pseudo-labels generated by the Teacher model. During the fine-tuning phase, we use the last $l = \left\lceil n/4 \right\rceil$ pseudo-labeled days as the validation set and the remaining $n-l$ days as the training set.

For both the Teacher and Student models, we used the Adam optimizer with a learning rate of 0.001 and mini-batch training with 64 samples. During fine-tuning, only the last convolutional layer and the fully-connected layers of the networks are trained for the classifiers that compose the Teacher model and for the MobileNet V3 Small Student model. Conversely, all layers are trained for the Custom Student models due to their smaller size. We train both the Teacher and Student models for 20 epochs. In both cases, we select the best-performing model based on validation data.

The images fed to the networks are in RGB format, sized at  $128\times 128$ for the Teacher models and the MobileNetV3 Small Student model, and $32\times 32$ for the Custom Student model. All reported results are an average of 5 runs.

\section{Experiments}\label{sec:experiments}

We adopt a cross-dataset scenario \cite{almeidaEtAl2022}, where one dataset serves as the training set for the Teacher models, and the other is used for testing and deploying the Student models. In Section \ref{sec:fixedDays}, we detail our experiments where the Student models are fine-tuned using pseudo-labeled data from seven days of the target dataset. In Section \ref{sec:numberOfDays}, we present experiments demonstrating the network improvements as we increase the number of pseudo-labeled days used in fine-tuning. Finally, in Section \ref{sec:trueLabels}, we present the results considering different \textit{a posteriori} values to get the pseudo-labels.

\subsection{One week of pseudo-labels}\label{sec:fixedDays}

In this section, we present the results from tests in which, for the initial $n=7$ days, the Teacher model classifies images from the target dataset. From the 8th day onward, a lightweight model fine-tuned with pseudo-labeled samples from the first 7 days is used for classification.

In Table \ref{table:resultsNDays}, we present the accuracy results on the test sets, excluding the initial 7 days to avoid bias. It shows the averaged accuracy of both the Custom and MobileNetV3 Small networks (Student models) after fine-tuning using the target camera angles. We also included the results if we continued using the Teacher models to classify the test instances after the 7th day, and the accuracy of the Student models before fine-tuning for comparison. The weighted average reported in Table \ref{table:resultsNDays} considers CNRPark-EXT and PKLot as test sets, accounting for the number of test samples in each dataset.

\begin{table}[htpb]
\caption{Results achieved considering $n=7$ days. Accuracy $\pm$ stdev.}
\centering
\setlength{\tabcolsep}{2.7pt}
%ExtraColStep para deixar separado os resultados da custom versus mobile
\begin{tabular}{@{\extracolsep{3pt}}rrrrr@{}}
\hline
\multicolumn{5}{l}{Train PKLot -  Test CNRPark-EXT} \\
% \multirow{2}{*}{Teacher} & \multicolumn{2}{c}{Custom Network} & \multicolumn{2}{c}{MobilenetV3 Small} \\
& \multicolumn{2}{c}{Custom Network} & \multicolumn{2}{c}{MobileNetV3 Small} \\\cline{2-3}\cline{4-5}
Teacher & \makecell{Without\\Fine-Tune} & Fine-Tuned & \makecell{Without\\Fine-Tune}& Fine-Tuned \\
$96.4\% \pm0.2$ & $90.0\% \pm1.0$ & $91.2\% \pm1.0$ & $95.6\% \pm0.6$ & $95.4\% \pm0.3$\\
\hline
\multicolumn{5}{l}{Train CNRPark-EXT -  Test PKLot} \\
% \multirow{2}{*}{Teacher} & \multicolumn{2}{c}{Custom Network} & \multicolumn{2}{c}{MobilenetV3 Small} \\
& \multicolumn{2}{c}{Custom Network} & \multicolumn{2}{c}{MobileNetV3 Small} \\\cline{2-3}\cline{4-5}
Teacher & \makecell{Without\\Fine-Tune} & Fine-Tuned & \makecell{Without\\Fine-Tune} & Fine-Tuned \\
$95.2\% \pm0.4$ & $80.2\% \pm2.0$ & $97.2\% \pm0.4$ & $91.3\% \pm1.0$ & $97.2\% \pm0.3$\\
\hline
\multicolumn{5}{l}{Weighted Average} \\
% \multirow{2}{*}{Teacher} & \multicolumn{2}{c}{Custom Network} & \multicolumn{2}{c}{MobilenetV3 Small} \\
& \multicolumn{2}{c}{Custom Network} & \multicolumn{2}{c}{MobileNetV3 Small} \\\cline{2-3}\cline{4-5}
Teacher & \makecell{Without\\Fine-Tune} & Fine-Tuned & \makecell{Without\\Fine-Tune} & Fine-Tuned \\
$95.3\% \pm0.4$ & $81.1\% \pm1.9$ & $96.6\% \pm0.5$ & $91.7\% \pm1.0$ & $97.0\% \pm0.3$\\
\hline
\end{tabular}
\label{table:resultsNDays}
\end{table}

As shown in Table \ref{table:resultsNDays}, the fine-tuned MobileNetV3 Small network achieved an accuracy of 0.4 percentage points higher on average than the Custom network. However, the MobileNetV3 Small network has approximately one order of magnitude more parameters than the Custom network. Therefore, the Custom network can better balance accuracy and computational efficiency under resources-constrained scenarios.

%It is known that training with pseudo-labels can exacerbate the biases of the model (e.g., any error committed by the Teacher model will be learned as a correct label by the Students).
Interestingly, on average, the Student models achieved better accuracies than the Teacher models, showing that it is possible to save computational power to run a lightweight model on the edge without sacrificing accuracy.
%This is true even for the Custom network, which has only 158,914 parameters, showing that it is possible to save computational power to run a lightweight model on the edge without sacrificing accuracy.
This phenomenon can be explained by the small number of parameters of the student networks and the relatively small number of wrongly classified instances by the Teacher model (see Section \ref{sec:trueLabels}), which may have helped to make the Student models specialized in the target scenario.

When considering the scenario where the Student models are fine-tuned and tested solely on the CNRPark-EXT dataset, there was only a small increase in the accuracy of the Custom network and a slight decrease in accuracy for the MobileNetV3 Small after the fine-tuning. This decrease can be attributed to the limited number of samples generated during the fine-tuning phase. In the CNRPark-EXT dataset, images are captured at intervals of 30 minutes from a narrow-angle, covering only a few parking spaces. For instance, considering camera 2 of CNRPark-EXT, on average, only 1,277 pseudo-labeled training samples were generated to fine-tune the models over the 7-day period examined in this section.

By a similar reasoning, during the tests using the PKLot, the accuracy was significantly increased after the fine-tuning (e.g., from 80.2\% to 97.2\% for the Custom model). First, in this scenario, the Student models are pre-trained using CNRPark-EXT images, which have a limited number of images and may have led to poor generalization. Second, with more open camera angles and one image taken every 5 minutes, a large number of pseudo-labeled images were generated in the PKLot to fine-tune the models. For example, considering the UFPR04 camera 20,884 pseudo-labeled samples were generated.

Finally, we tested the system's deployment on a Raspberry Pi 5 to check the time taken to classify parking spots on a portable device using the Custom network. We implemented the system using Python, OpenCV, and Pytorch without considering any optimization. We assumed that the camera's collected image was already saved in the device's permanent memory, and we computed the time taken for the complete process, from loading and cropping the image to classifying its parking spaces. On average, it took 0.01 seconds to process each spot as occupied/empty. If we consider that a camera can cover 100 parking spaces in its field of view, it will take 1 second to refresh the statuses of all visible parking spaces using the Raspberry Pi 5. It is important to notice that this result is not directly comparable to the state-of-the-art since we are using the most recent version of the Raspberry Pi hardware.

\subsection{Pseudo-Labeled Days Versus Accuracy}\label{sec:numberOfDays}

In this Section, we examine the performance of the Student models using varying numbers of days $n \in [6..14]$ for image collection for fine-tuning. We aim to analyze whether increasing the amount of pseudo-labeled data for fine-tuning improves the accuracy of the Student models.

% We tested values $n \in [6..14]$. These values were chosen since 6 is the minimum number of days we can use that generates instances from both classes in the training and validation sets, considering all camera angles available in the PKLot and CNRPar-EXT datasets. Also, as the CNRPark-EXT contains only 23 days of data collection, by considering 14 days, there are 9 days remaining to test the models.

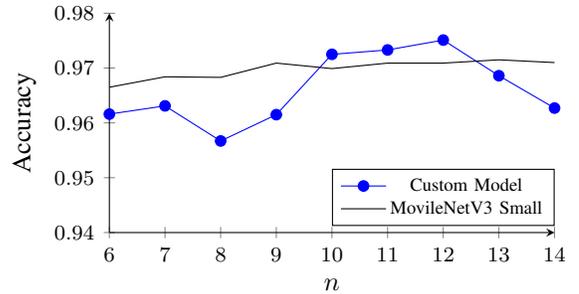
\begin{figure}[!htbp]
    \centering
    \begin{tikzpicture}
    \begin{axis}[
        width=7.5cm,
        height=4.5cm,
        ymin=0.94,
        ymax=0.98,
        ticklabel style = {font=\footnotesize},
        axis x line=bottom,
        axis y line=left,
        xtick distance=1,
        %ytick distance=500,
        scaled y ticks=false,
        xlabel=$n$,
        ylabel=Accuracy,
        legend style={nodes={scale=0.7, transform shape}},
        legend style={at={(0.50,0.03)},anchor=south west}
    ]
     \addplot [color=blue,mark=*] table [x=day,y=acccust, col sep=comma]  {tikz/trainCnrTestPklot.csv};
     \addlegendentry{Custom Model};
     % \addplot [color=blue,] table [x=day,y=acccustpre, col sep=comma]  {tikz/trainCnrTestPklot.csv};
     % \addlegendentry{Ground Truth};
     \addplot [color=black] table [x=day,y=accmnv3, col sep=comma]  {tikz/trainCnrTestPklot.csv};
     \addlegendentry{MovileNetV3 Small};
     % \addplot [color=black] table [x=day,y=accmnv3pre, col sep=comma]  {tikz/trainCnrTestPklot.csv};
     % \addlegendentry{Prediction};
    \end{axis}
\end{tikzpicture}
    \caption{Results varying the number of days to fine-tune the models $n$ using the PKLot as the test set.}
  \label{fig:nTestPklot}
\end{figure}

In the results presented, we exclude the first $14$ days of the test sets to make the results comparable. Figures \ref{fig:nTestPklot} and \ref{fig:nTestCNR} show the results considering the PKLot and CNRPark-EXT as test sets, respectively. As one can observe, the bigger the value of $n$, the better the result, especially when considering the custom model. These results indicate that, when feasible, it is worth keeping the system depending on the Teacher models for longer times.

\begin{figure}[!htbp]
    \centering
    \begin{tikzpicture}
    \begin{axis}[
        width=7.5cm,
        height=4.5cm,
        ymin=0.88,
        ymax=0.96,
        ticklabel style = {font=\footnotesize},
        axis x line=bottom,
        axis y line=left,
        xtick distance=1,
        %ytick distance=500,
        scaled y ticks=false,
        xlabel=$n$,
        ylabel=Accuracy,
        legend style={nodes={scale=0.7, transform shape}},
        legend style={at={(0.50,0.03)},anchor=south west}
    ]
    \addplot [color=blue,mark=*] table [x=day,y=acccust, col sep=comma]  {tikz/trainPklotTestCNR.csv};
    \addlegendentry{Custom Model};
    \addplot [color=black] table [x=day,y=accmnv3, col sep=comma]  {tikz/trainPklotTestCNR.csv};
    \addlegendentry{MovileNetV3 Small};
    \end{axis}
\end{tikzpicture}
    \caption{Results varying the number of days to fine-tune the models $n$ using the CRNPark-EXT as the test set.}
  \label{fig:nTestCNR}
\end{figure}

\subsection{A Posteriori thresholds versus Accuracy}\label{sec:trueLabels}

In this Section, we use a similar procedure used in Section \ref{sec:fixedDays}, where for the initial $n=7$ days, the Teacher model classifies the images, and from the 8th day onward, a lightweight model trained with the pseudo-labeled samples is deployed. Nevertheless, in this Section, we vary the \textit{a posteriori} threshold used to select the pseudo-labeled samples. We select only images classified with \textit{a posteriori} probabilities (given by the Teacher model) greater than $\in [0.5,0.6,0.7,0.8,0.9]$.

We used only the Custom Network. The results are shown in Table \ref{table:aPosterioriThresh}, where the accuracy refers to the accuracy achieved after the student model's fine-tuning in the test set. The Used column refer to the total number of samples (summing all camera angles available in each test set) above the minimum \textit{a posteriori} probability necessary, and thus that were used as pseudo-labels, and the Wrong column shows the total number of used pseudo-labeled samples with wrong labels. The last line of Table \ref{table:aPosterioriThresh} shows the result of a hypothetical oracle capable of generating perfect pseudo labels for the test instances. The oracle serves as a ceiling for the results, showing the best result achievable.

\begin{table}[htpb]
\caption{Results achieved considering different \textit{a posteriori} thresholds and the true labels of the target dataset.}
\centering
\setlength{\tabcolsep}{1.5pt}
\begin{tabular}{@{\extracolsep{5pt}}rrrrrrr@{}}\hline
 & \multicolumn{3}{c}{PKLot as test set} & \multicolumn{3}{c}{CNRPark-EXT as test set} \\\cline{2-4}\cline{5-7}
\textit{a p}. & Accuracy & Used & Wrong &  Accuracy & Used & Wrong\\\hline
0.5 & $94.7\% \pm0.7$ & 225,279 & 10,012 & $90.1\% \pm0.7$ & 45,467 & 1,276\\
0.6 & $94.9\% \pm0.7$ & 214,071 & 5,735 & $90.3\% \pm0.9$ & 44,667 & 934\\
0.7 & $95.9\% \pm0.6$ & 200,089 & 2,931 & $91.0\% \pm0.6$ & 43,591 & 627\\
0.8 & $96.4\% \pm0.6$ & 179,248 & 1,181 & $91.1\% \pm0.6$ & 41,915 & 363\\
0.9 & $97.2\% \pm0.4$ & 141,209 & 312 &  $91.2\% \pm1.0$ & 38,453 & 168\\\hline
True & $97.2\% \pm0.4$ & 225,782 & - & $92.8\% \pm0.6$ & 45,493 & -\\\hline
\end{tabular}
\label{table:aPosterioriThresh}
\end{table}

The results in Table \ref{table:aPosterioriThresh} show that, as expected, the greater the \textit{a posteriori} threshold, the less wrong pseudo-labeled samples are generated to train the Student models, with the cost of generating less usable samples to train the models. The results also show that using pseudo-labeled samples makes it possible to achieve results similar to those achieved by using true labels. The results using an a posteriori threshold of 0.9 reached virtually the same results of the models fine-tuned using the true labels considering the PKLot test set, and the results were 1.6\ percentage points bellow the models trained with the true labels considering the CNRPark-EXT test set. This behavior can be explained by the relatively small number of wrongly labeled samples generated by the Teacher models when considering the 0.9 threshold (e.g., only 312, or 0,22\%, of the 141,209 images pseudo-labeled in the PKLot test set).

\section{Conclusion}\label{sec:conclusion}

In this work, we propose using an ensemble of classifiers as Teacher models to generate pseudo-labeled samples to fine-tune lightweight Student models for the parking spaces classification problem. The Student models are lightweight and fine-tuned using specific pseudo-labeled data from the environment where they will be deployed. The Teacher ensembles consist of MobileNetV3 Large networks, and the Student models are the MobileNetV3 Small and a Custom 3-layer network. We reach the following conclusions for our research questions:

\textbf{RQ1: How does the accuracy of the lightweight Student classifiers compare to the central Teacher model?} The Custom 3-layer network, which has 26 times fewer parameters than the networks used in the Teacher ensemble, was able to reach better results than the Teacher ensemble after the fine-tunning using the pseudo-labels, showing that lightweight models, that can be deployed on devices with restricted computing power, can reach good results without the need of manual labeling of data.

\textbf{RQ2: How many images must be classified by the Teacher model to create enough pseudo-labels?} We show that, on average, seven days of pseudo-labeled are enough to fine-tune the Student models and reach better accuracies than the Teacher models. The experiments also show that the more time is spent on pseudo-labeling days with the Teacher models, the better the result of the fine-tuning of the Student models.

\textbf{RQ3: How does the accuracy of the Student models fine-tuned with the pseudo-labels compare with hypothetical Students fine-tuned with true-labels?} In our experiments, we show that, on average, only 0,27\% of pseudo-labeled samples were wrongly labeled by the Teacher ensemble considering an \textit{a posteriori} threshold of 0.9. Thus, the fine-tuned Student reached an accuracy only 1.6 percentage points smaller than the model fine-tuned with the true labels when considering the CNRPar-EXT test set, which was the worst-case scenario.

Thus, our experiments show that it is possible to use computationally expensive Teacher models to fine-tune lightweight Students for parking space classification. This makes it feasible to deploy such systems on a large scale in smart cities, for instance. In future work, we plan to extend the idea to other smart-city problems, such as traffic and pedestrian monitoring.

%\balance
\bibliographystyle{IEEEtran}
\bibliography{IEEEabrv,main}

\end{document}